\documentclass[twoside]{article}

\usepackage[accepted]{aistats2019}

\usepackage{amsfonts,amsmath,amsthm}
\usepackage{graphicx} 
\usepackage{color}
\usepackage{algorithm, algorithmic,cases}
\usepackage[mathscr]{euscript}

\usepackage{multirow}
\usepackage{subcaption}

\usepackage{xfrac}
\usepackage[nice]{nicefrac}

\usepackage[utf8]{inputenc} 
\usepackage[T1]{fontenc}    
\usepackage{hyperref}       
\usepackage{url}            
\usepackage{booktabs}       
\usepackage{amsfonts}       
\usepackage{nicefrac}       
\usepackage{microtype}      

\usepackage{comment}

\newcommand{\po}{P^o}
\newcommand{\bopar}{\bo}
\newcommand{\paug}[1]{P^{(#1)}}

\newcommand{\ppio}{p^{\pi^{o}}}

\newcommand{\dmu}[1]{d^\mu(#1|o)}
\renewcommand{\O}{\mathscr{O}}
\newcommand{\margP}{\po_{\mu}}

\newcommand{\M}{\mathscr{M}} 
\newcommand{\A}{\mathscr{A}} 
\newcommand{\X}{\mathscr{X}} 

\newcommand{\E}{\mathbb{E}} 
\newcommand{\Exp}[2]{\E_{#1}\left[#2\right]}
\newcommand{\1}{\mathbb{I} } 
\newcommand{\indic}[1]{\1_{#1}} 
\newcommand\defequal{\mathrel{\overset{\makebox[0pt]{\mbox{\tiny def}}}{=}}}

\newcommand{\pio}{{\pi^o}} 
\newcommand{\bo}{\beta^o}
\newcommand{\I}{\mathscr{I}} 
\newcommand{\Io}{\I^o}

\newcommand{\Natural}{\mathbb{N}}
\newcommand{\bN}{\Natural}

\newtheorem{theorem}{Theorem}
\newtheorem{corollary}{Corollary}
\newtheorem{proposition}{Proposition}
\newtheorem{assumption}{Assumption}

\allowdisplaybreaks

\begin{document}

\twocolumn[

\aistatstitle{The Termination Critic}

\aistatsauthor{ Anna Harutyunyan \And Will Dabney \And  Diana Borsa} 

\aistatsaddress{ DeepMind \And DeepMind \And DeepMind}

\aistatsauthor{ Nicolas Heess \And R\'{e}mi Munos \And Doina Precup }

\aistatsaddress{DeepMind \And DeepMind \And DeepMind } ]

\runningauthor{Anna Harutyunyan, Will Dabney, Diana Borsa, Nicolas Heess, R\'{e}mi Munos, Doina Precup}

\begin{abstract}
In this work, we consider the problem of autonomously discovering behavioral abstractions, or options, for reinforcement learning agents. We propose an algorithm that focuses on the {\em termination condition}, as opposed to -- as is common -- the policy. The termination condition is usually trained to optimize a control objective: an option ought to terminate if another has better value. We offer a different, information-theoretic perspective, and propose that terminations should focus instead on the {\em compressibility} of the option's encoding -- arguably a key reason for using abstractions.
To achieve this algorithmically, we leverage the classical options framework, and learn the option transition model as a “critic” for the termination condition.
Using this model, we derive gradients that optimize the desired criteria. 
We show that the resulting options are non-trivial, intuitively meaningful, and useful for learning and planning.
\end{abstract}

\section{Introduction}

Autonomous discovery of meaningful behavioral abstractions in reinforcement learning has proven to be surprisingly elusive. One part of the difficulty perhaps is the fundamental question of {\em why} such abstractions are needed, or useful. Indeed, it is often the case that a primitive action policy is sufficient, and the overhead of discovery outweighs the potential speedup advantages. In this work, we adopt the view that abstractions are primarily useful due to their ability to {\em compress} information, which yields speedups in learning and planning. For example, it has been argued in the neuroscience literature that behavior hierarchies are optimal if they induce plans of minimum description length across a set of tasks~\cite{solway2014optimal}.

 Despite significant interest in the discovery problem, 
 only in recent years have there been algorithms that tackle it with minimal supervision. An important example is the option-critic~\cite{bacon2017option}, which focuses on options~\cite{sutton1999between} as the formal framework of temporal abstraction and learns both of the key components of an option, the policy and the termination, end-to-end.
 Unfortunately, in later stages of training, the options tend to collapse to single-action primitives. This is in part due to using the advantage function as a training objective of the termination condition: an option ought to terminate if another option has better value. This, however, occurs often throughout learning, and can be simply due to noise in value estimation. 
 The follow-up work on option-critic {\em with deliberation cost}~\cite{harb2017waiting} addresses this option collapse by modifying the termination objective to additionally penalize option termination, 
 but it is highly sensitive to the associated cost parameter.
 
 In this work, we take the idea of modifying the termination objective to the extreme, and propose for it to be completely independent of the task reward. Taking the compression perspective, we suggest that this objective should be information-theoretic, and should capture the intuition that it would be useful to have "simple" option encodings that focus termination on a small set of states. We show that such an objective correlates with the planning performance of options for a set of goal-directed tasks.
 

Our key technical contribution is the manner in which the objective is optimized. We derive a result that relates the gradient of the option transition model to the gradient of the termination condition, allowing one to express objectives in terms of the option model, and optimize them directly via the termination condition. The model hence acts as a "critic", in the sense that it measures the quality of the termination condition in the context of the objective, analogous to how the value function measures the quality of the policy in the context of reward. Using this result, we obtain a novel policy-gradient-style 
algorithm which learns terminations to optimize our objective, and policies to optimize the reward as usual. The separation of concerns is appealing, since it bypasses the need for sensitive trade-off parameters. We show that the resulting options are non-trivial,  and useful for learning and planning. 

The paper is organized as follows. After introducing relevant background, we present the termination gradient theorem, which relates the change in the option model to the change in terminations. We then formalize the proposed objective, express it via the option model, and use the termination gradient result to obtain the 
online actor-critic termination-critic (ACTC) algorithm. Finally, we empirically study the learning dynamics of our algorithm, the relationship of the proposed loss to planning performance, and analyze the resulting options qualitatively and quantitatvely. 


\section{Background and Notation}
We assume the standard reinforcement learning (RL)
setting~\cite{sutton-barto17} of a Markov Decision Process (MDP) $\M =
(\X, \A, p, r, \gamma)$, where $\X$ is the set of states, $\A$ the set
of discrete actions; $p:\X\times\A\times \X \rightarrow [0,1]$ is  the
transition model that specifies the environment dynamics, with $p(x'|x, a)$
denoting the probability of transitioning to state $x'$ upon taking
action $a$ in $x$; $r : \X\times\A \rightarrow
[-r_{\max},r_{\max}]$ the reward function, and $\gamma\in [0,1]$ the
scalar discount factor. A policy is a probabilistic mapping from
states to actions. For a policy $\pi$, let the matrix $p^\pi$ denote the
dynamics of the induced Markov chain: $p^\pi(x'|x) = \sum_{a\in\A}\pi(a|x)p(x'|x,a),$ and $r^\pi$ the reward expected for each state under $\pi$: $r^\pi(x) = \sum_{a\in\A} \pi(a|x) r(x,a).$

The goal of an RL agent is to find a policy $\pi$ that produces the highest expected cumulative reward:
\begin{equation}
    J(\pi) = \Exp{}{\sum_{t=1}^\infty \gamma^{t-1} R_{t} | x_0, \pi }, \label{eq:J-pi}
\end{equation}
where $R_t=r(X_t,A_t)$ is the random variable corresponding to the reward received at time $t$, and $x_0$ is the initial state of the process. An {\em optimal} policy is one that maximizes $J$. A related instrumental quantity in RL is the action-value (or Q-) function, which measures $J$ for a particular state-action pair:
\begin{equation} 
q^\pi(x, a) = \Exp{}{\sum_{k=1}^\infty \gamma^{k-1} R_{t+k} | X_t = x, A_t = a, \pi}. \label{eq:q}
\end{equation}


The simplest way to optimize the objective~\eqref{eq:J-pi} is directly by adjusting the policy parameters $\theta_{\pi}$ (assuming $\pi$ is differentiable)~\cite{williams1992simple}. The {\em policy gradient (PG)} theorem~\cite{sutton2000policy} states that:
\begin{equation}
    \nabla_{\theta_\pi} J(\pi) = \sum_{x} d^\pi(x) \sum_a \nabla_{\theta_\pi} \pi(x, a) q^\pi(x, a) \notag
\end{equation}
where $d_{\pi}$ is the stationary distribution induced by $\pi$. Hence, samples of the form $\nabla_{\theta_\pi} \log \pi(x, a) q^\pi(x, a)$ produce an unbiased estimate of the gradient $\nabla_{\theta_\pi} J(\pi)$.

The final remaining question is how to estimate $q^\pi$. A standard answer relies on the idea of {\em temporal-difference (TD) learning}, which itself leverages sampling of the following recursive form of Eq.~\eqref{eq:q}, known as the Bellman Equation~\cite{bellman1957dynamic}:
\begin{equation} 
q^\pi(x, a) = r(x, a) + \gamma \sum_{x'} p(x'|x,a) \sum_{a'} \pi(a'|x')  q^{\pi}(x',a'). \notag 
\end{equation}
Iterating this equation from an arbitrary initial function $q$ produces the correct $q^\pi$ in expectation (e.g.~\cite{puterman94}).




\subsection{The Options Framework}
Options provide the standard formal framework for modeling temporal abstraction in RL~\cite{sutton1999between}. An option $o$ is a tuple $(\Io, \bo, \pio)$. Here, $\Io\subseteq\X$ is the
initiation set, from which option $o$ may start (as in other recent work, we take $\Io=\X$ for simplicity),
$\bo:\X\rightarrow
[0,1]$ is the termination condition, with $\bo(x)$ 
denoting the probability of option $o$ terminating in state $x$; and $\pio$ is
the internal policy of option $o$. As is common, we assume that options eventually terminate:
 \begin{assumption}
\label{ass:reachable}
  For all $o\in\O$ and $x\in\X$, $\exists x_f$ that is reachable by $\pio$ from $x$, s.t. $\bo(x_f) > 0$.
\end{assumption}

Analogously to the one-step MDP reward and transition models $r$ and $p$, options 
induce {\em semi-}MDP~\cite{puterman94} reward and transition models:
 \begin{align*}
 \po(x_f|x_s) & = \gamma\bo(x_f)p^{\pio}(x_f|x_s) \\
 & \qquad + \gamma\sum_{x}p^{\pio}(x|x_s) (1-\bo(x)) P^o(x_f|x), \\
 R^o(x_s) & = r^{\pio}(x_s) + \gamma\sum_{x}p^{\pio}(x|x_s) (1-\bo(x)) R^o(x).
 \end{align*}
 That is: the option transition model $P^o(\cdot|x_s)$
 outputs a sub-probability distribution over states, which, for each $x_f$ captures the discounted probability of reaching $x_f$ from $x_s$ (in any number of steps) {\em and} terminating there.
In this work, unless otherwise stated, we will use a slightly different formulation, which is undiscounted and shifted backwards by one step:
\begin{align}
\po (x_f|x_s) & = \bo(x_f)\indic{x_f=x_s} \notag 
\\ &  + (1-\bo(x_s)) 
\sum_{x}p^{\pio}(x|x_s)\po(x_f|x),
\label{eq:po}
\end{align} 
where $\indic{}$ denotes the indicator function. The lack of discounting is convenient because it leads to $\po$ being a probability (rather than sub-probability) distribution over $x_f$ so long as Assumption~\ref{ass:reachable} holds.\footnote{In particular, if we sum over $x_f$ we obtain the Poisson binomial (or sequential independent Bernoulli) probability of one success out of infinite trials, which is equal to 1.}
The backwards time shifting replaces the first $\bo(x_f)\ppio(x_f|x_s)$ term with $\bo(x_f)\indic{x_f=x_s} = \bo(x_s)\indic{x_f=x_s}$, which conveniently considers $\bo$ of the same state, and will be important for our derivations.

We will use $\mu$ to denote the policy over options, and define the option-level Q-function as in~\cite{sutton1999between}.

\newcommand{\kl}{D_{KL}}


\section{The Termination Gradient Theorem} 
\label{sec:terminate}


The starting point for our work is the idea of optimizing the option's termination condition independently and with respect to a different objective than the policy. How can one easily formulate such objectives? We propose to leverage the relationship of the termination condition to the option model $\po$, which is a distribution over the final states of the option, a quantity that is relevant to many natural objectives. The classical idea of terminating in "bottleneck" states is an example of such an objective.

To this end, we first note that the 
definition~\eqref{eq:po} of $\po$ has a formal similarity to a Bellman equation, in which  $\po$ is akin to a value function and $\bo$ influences both the immediate reward {\em and} the discount. Hence, we can
express the gradient of $\po$ through the gradient of $\bo$. This will allow us to express high-level objectives through $\po$ and optimize them through $\bo$. The following general result formalizes this relationship, while the next section uses it for a particular objective. The theorem concerns a single option $o$, and we write $\theta_\beta$ to denote the parameters of $\bo$.


\newcommand{\deriv}{\nabla_{\theta_{\beta}}}
\newcommand{\dP}[1]{\deriv\po{(#1)}} 
\newcommand{\dbeta}[1]{\deriv\bopar{(#1)}} 

\newcommand{\rox}[2]{r^o_{#2}(#1)}
\newcommand{\dox}[2]{d^o_{#1}(#2)}
\newcommand{\dlogbeta}[1]{\deriv\log\bo(#1)}

\begin{theorem}
\label{thm:po-gradient}
Let $\bo(x) \in (0, 1), \forall x\in\X, o\in\O$ be parameterized by $\theta_{\beta}$. The gradient of the option model w.r.t. $\theta_{\beta}$ is:
\begin{equation}
    \dP{x_f|x_s} = \sum_{x} \po(x|x_s) \dlogbeta{x}\rox{x}{x_f}, \notag 
\end{equation}
where the "reward" term is given by:
\begin{equation}
    \rox{x}{x_f} \defequal \frac{\indic{x_f=x} - \po(x_f|x)} {1-\bo(x)} \notag 
\end{equation}
\end{theorem}
The proof is given in appendix. In the following, we will drop the superscript and simply write $\theta_\beta$. Note that in the particular (common) case when $\beta$ is parameterized with a sigmoid function, the expression takes the following simple shape:
\renewcommand{\deriv}{\nabla_{\theta_{\beta}}}
\begin{equation}
    \dP{x_f|x_s} = \sum_{x} \po(x|x_s) \deriv\ell_{\bo}(x)(\indic{x_f=x} - \po(x_f|x)), \notag 
\end{equation}
where $\ell_{\bo}(x)$ denotes the logit of $\bo(x)$. The pseudo-reward $\indic{x_f=x} - \po(x_f|x)$ has an intuitive interpretation. In general, it is always positive at $x_f$ (and so $\bo(x_f)$ should always increase to maximize $\po(x_f|x_s)$), and negative at all other states $x\neq x_f$ (and so $\bo(x)$ should always decrease to maximize $\po(x_f|x_s)$). The amount of the change depends on the dynamics $\po(x_f|x)$. 
\begin{itemize}
    \item In terminating states $x_f$, if $\po(x_f|x_f)$ -- the likelihood of terminating in $x_f$ immediately {\em or} upon a later return -- is low, then the change in $\bo(x_f)$ is high: an immediate termination needs to occur in order for $\po(x_f|x_s)$ to be high. If $\po(x_f|x_f)$ is high, then $\po(x_f|x_s)$ may be high without $\bo$ needing to be high.\footnote{E.g. if there is a single terminating state that is guaranteed to be reached, any $\bo(x_f) > 0$ will do.}
    \item In non-terminating states $x \neq x_f$, the higher $\po(x_f|x)$, or the likelier it is for the desired terminating state $x_f$ to be reached from $x$, the more the termination probability at $x$ is reduced, to ensure that this happens. If $x_f$ is not reachable from $x$ no change occurs at $x$ to maximize $\po(x_f|x_s)$.
\end{itemize}

The theorem can be used as a general tool to optimize any differentiable objective of $\po$. In the next section we propose and justify one such objective, and derive the complete algorithm.

\section{The Termination Critic}



We now propose the specific idea for the actor-critic termination critic (ACTC) algorithm. We first formulate our objective, then use Theorem~\ref{thm:po-gradient} to derive the gradient of this objective as a function of the gradient of $\bo$. Finally, we formulate the online algorithm that estimates all of the necessary quantities from samples.

\subsection{Predictability Objective}

We postulate that desirable options are "targeted" and have small terminating regions.
This can be expressed as having a low entropy distribution of final states. We hence propose to explicitly minimize this entropy as an objective:
\begin{equation}
    J(P^o) = H(X_f | o)
    \label{eq:criteria-generic}
\end{equation}
where $H$ denotes entropy, and $X_f$ is the random variable denoting a terminating state. We call this objective {\em predictability}, as it measures how predictable an option's outcome (final state) is.
Note that the entropy is marginalized over all starting states. The marginalization allows for {\em consistency} -- without it, the objective can be satisfied for each starting state individually without requiring the terminating states to be the same. We will later show that this objective indeed correlates with planning performance (Section~\ref{sec:correlation-with-planning}).

In the exact, discrete setting the objective~\eqref{eq:criteria-generic} is minimized when an option terminates at a single state $x_f$. Note that the this is the case irrespective of the choice of $x_f$. The resulting $x_f$ will hence be determined by the learning dynamics of the particular algorithm. We will show later empirically that the gradient algorithm we derive is attracted to $x_f$-s that are most visited, as measured by $\po$ (Section~\ref{sec:learning-dynamics}).


The objective is reminiscent of other recent information-theoretic option-discovery approaches (e.g.~\cite{gregor2016variational,florensa2017stochastic,hausman2018learning}), but is focused on the shape of each option in isolation. Similar to those works, one may wish to explicitly optimize for {\em diversity} between options as well, so as to encourage specialization. This can be attained for example by maximizing the complete mutual information $I(X_f|O) = H(X_f) - H(X_f|O)$ as the objective (rather than only its second term). In this work we choose to keep the objective minimalistic, but will observe some diversity occur under a set of tasks, due to the sampling procedure, so long as the policy over options is non-trivial. 

The manner in which we propose to optimize this objective is novel and entirely different from existing work. We leverage the analytical gradient relationship derived in Theorem~\ref{thm:po-gradient}, and so instead of estimating the entropy term directly, we will express it through the option model, and estimate this option model. The following proposition expresses criteria~\eqref{eq:criteria-generic} via the option model. 
\begin{proposition}
\label{prop:criteria-po}
Let $\mu$ denote the policy over options, and let $\dmu{\cdot}$ be option $o$'s starting distribution induced by $\mu$. Let $\margP(x_f) \defequal \sum_{y_s}\dmu{y_s}\po(x_f|y_s).$  The criteria~\eqref{eq:criteria-generic} 
can be expressed via the option model as follows: 
\begin{equation}
    J(P^o) = -\E_{x_s} \Big[\sum_{x_f} \po(x_f|x_s)\log\margP(x_f) \Big] \notag
    \end{equation}
\end{proposition}
The proof is mainly notational and given in appendix.  This proposition implies that for a particular start state $x_s$, our global entropy loss can be interpreted as a cross-entropy loss between the distributions $\po(\cdot|x_s)$ of final states from a particular state $x_s$ and $\margP(\cdot)$ of final states from all start states.

\paragraph{A note on Assumption~\ref{ass:reachable}.} Finally, we note that the entropy expression is only meaningful if $\po$ is a probability distribution, otherwise it is for example minimized by an all-zero $\po$. In turn, $\po$ is a distribution if Assumption~\ref{ass:reachable} holds, but if the option components are being learned, this may be tricky to uphold, and a trivial solution may be attractive. We will see empirically that the algorithm we derive does not collapse to this trivial solution. However, a general way of ensuring adherence to Assumption~\ref{ass:reachable} 
remains an open problem. 
 
\subsection{The Predictability Gradient}
\newcommand{\tb}{\theta_\beta}
\newcommand{\termcorr}{\omega_{x_f}(x')} 


We are now ready to express the gradient of the overall objective from Proposition~\ref{prop:criteria-po} in terms of the termination parameters $\tb$. This result is analogous to the policy gradient theorem, and similarly, for the gradient to be easy to sample, we need for a certain distribution to be independent of the termination condition.

\begin{assumption}
\label{ass:independence}
The distribution $\dmu{\cdot}$ over the starting states of an option $o$ under policy $\mu$ is independent of its termination condition $\bo$.
\end{assumption}

This is satisfied for example by any fixed policy $\mu$. Of course, in general $\mu$ often depends on the value function over options $q$ which in turn depends on $\bo$, but one may for example apply two-timescale optimization to make $q$ appear quasi-fixed (e.g.~\cite{borkar1997stochastic}). Our experiments show that even when not doing this explicitly, the online algorithm converges reliably. The following theorem derives the gradient of the objective~\eqref{eq:criteria-generic} in terms of the gradient of $\bo$.


\newcommand{\corr}[1]{C^{o'}_{#1}(x')}
\newcommand{\bop}{\beta^{o'}}
\begin{theorem}
\label{thm:main-result}
 Let Assumption~\ref{ass:independence} hold. We have that:
\begin{align}
    \deriv J(\po) & = -\sum_{x_s} \dmu{x_s} \sum_{x} \frac{\po(x|x_s)}{\bo(x)} \frac{\dbeta{x}}{1-\bo(x)} \times \notag \\
    & \quad \Bigg[ \underbrace{\log \margP(x) - \sum_{x_f} \po(x_f|x) \log \margP(x_f)}_{\mbox{\small reachability advantage $A^o_P(x)$}}  \notag \\
    & \quad + \underbrace{1 - \sum_{x_f} \po(x_f|x) \frac{\po(x_f|x_s)\margP(x)}{\margP(x_f)\po(x|x_s)}}_{\mbox{\small trajectory advantage $A^o_\tau(x | x_s)$}} \Bigg] \notag
\end{align}
\end{theorem}
The proof is a fairly straightforward differentiation, and is given in appendix. The loss consists of two terms. The {\em termination advantage} measures how likely option $o$ is to reach and terminate in state $x$, as compared to other alternatives $x_f$. The {\em trajectory advantage} measures the desirability of state $x$ in context of the starting state $x_s$ -- if $x$ is a likely termination state {\em in general}, but not for the particular $x_s$, this term will account for it. Appendix~\ref{app:learning-dynamics} studies the effects of these two terms on learning dynamics in isolation.

Now, we would like to derive a sample-based algorithm that produces unbiased estimates of the derived gradient. Substituting the expression for the gradient of the logits, we can rewrite the overall expression as follows:
\begin{align*}
    \deriv J(\po) & = -\E_{x_s} \E_{x} \deriv\ell_{\bo}(x) \Big(  A^o_P(x) + A^o_\tau(x | x_s)  \Big)
\end{align*}
where the two expectations are w.r.t. the distributions written out in Theorem~\ref{thm:main-result}. We will sample these expectations from  trajectories of the form $\tau = x_s, \ldots, x, \ldots, x_f$, and base our updates on the following corollary.
\begin{corollary}
\label{cor:update}
Consider a sample trajectory $\tau = x_s, \ldots, x, \ldots, x_f$, and let $\bo$ be parameterized by a sigmoid funciton. For a state $x$,
the following gradient is an unbiased estimate of $\deriv J(\po)$:
\begin{align}
- \deriv \ell_{\bo}(x) \bo(x) \Big( & \tilde{A}^o_P(x, x_f) + \tilde{A}^o_\tau(x, x_f | x_s)\Big), \label{eq:update-x}\\
 \tilde{A}^o_P(x, x_f) & =  \log \margP(x) - \log \margP(x_f) \notag \\
 \tilde{A}^o_\tau(x, x_f | x_s) & = 1 - \frac{\po(x_f|x_s)\margP(x)}{\margP(x_f)\po(x|x_s)} \notag 
\end{align}
where $\ell_{\bo}(x)$ denotes the logit of $\bo$ at state $x$, and $\tilde{A}^o_P(x, x_f)$, $\tilde{A}^o_\tau(x, x_f|x_s)$ are samples from the corresponding advantages for a particular $x_f$.
\end{corollary}
The $\bo(x)$ factor in Eq.~\eqref{eq:update-x} is akin to an importance correction necessary due to not actually having terminated at $x$ (and hence not having sampled $\po(x|x_s)$). Note that the state $x_f$ itself never gets updated directly, because the sampled advantages for it are zero (although the underlying state still may get updates when not sampled as final). The resulting magnitude of termination values at chosen final states hence depends on the initialization. To remove the dependence, we will deploy a baseline in the complete algorithm.

\section{Algorithm}

We now give our algorithm based on Corollary~\ref{cor:update}. In order to do so, we need to simultaneously estimate the transition model $\po$. Because for a given final state, it is simply a value function, we can readily do this with temporal difference learning. Furthermore, we need to estimate $\margP(x_f)$, which we do simply as an empirical average of $\po(x_f|x_s)$ over all experienced $x_s$. Finally, we deploy a per-option baseline $B^o$ that tracks the empirical average update and gets subtracted from the updates at all states. The complete end-to-end algorithm is summarized in Algorithm~\ref{alg:actc}. Similarly to e.g. A3C~\cite{mnih2016asynchronous}, our algorithm is online up to the trajectory length. One can potentially be fully online by instead of relying on the sampled terminating state $x_f$, "bootstrapping" with the value of $\po(x_f|x)$, but this requires an accurate estimate of $\po$.


\renewcommand{\O}{\mathscr{O}} 


\begin{algorithm}[t]
\caption{Actor-critic termination critic (ACTC)}
\label{alg:actc}
\begin{algorithmic}
\medskip
\item[\textbf{Given:}] Initial Q-function $q_0$, step-sizes
  $(\alpha_k)_{k \in \bN}$
\FOR{$k = 1, \ldots$}
    \STATE{Sample a trajectory $x_s, \dots, x_i, \dots, x_f$ for $T_k$ steps}
    \vspace{-1em}
    \FOR{all $x_i$ in the trajectory}
        \STATE Update $\beta(x_i)$ with TG via \eqref{eq:update-x}
        \STATE Update $\po(x_f|x_{i})$ with TD via \eqref{eq:po}
        \STATE Update $\margP(x_f)$ with MSE towards $\po(x_f|x_i)$
        \STATE Update $B^o$ with MSE towards \eqref{eq:update-x} 
        \STATE Update $\pi^o(a_i|x_i)$ with PG via the multi-step advantage w.r.t. $Q$
        \STATE Update $Q(x_i,o)$ with TD in a standard way 
    \ENDFOR
\ENDFOR
\end{algorithmic}
\end{algorithm}

 \begin{figure}[b]
     \centering
     \includegraphics[scale=0.4]{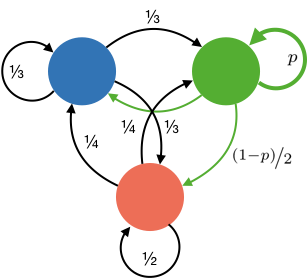}
     \caption{Example MDP} 
     \label{fig:toy-mdp}
 \end{figure}

 \begin{figure*}
     \centering
     \includegraphics[scale=0.37]{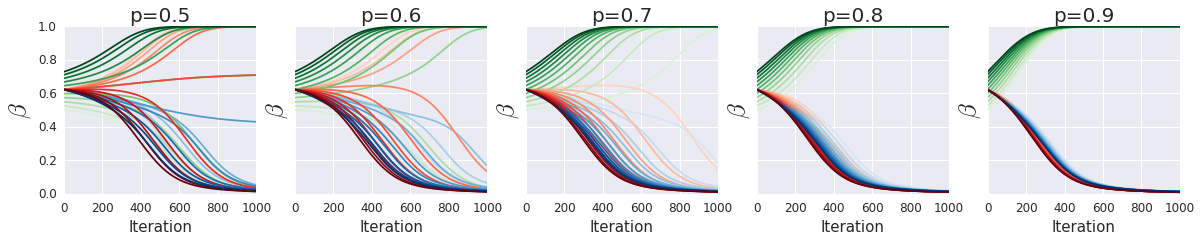}
     \includegraphics[scale=0.37]{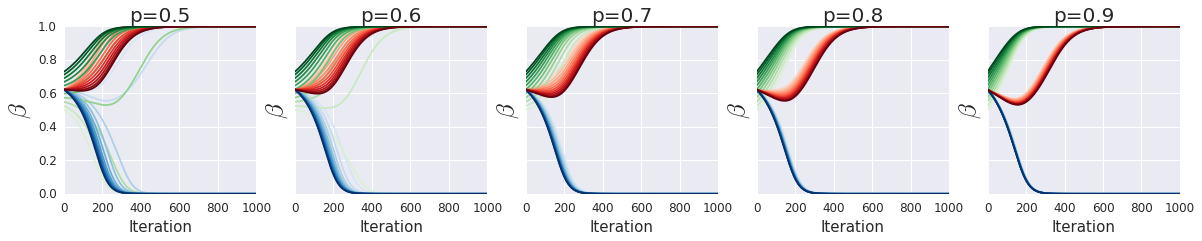}
     \caption{Learning dynamics. The color groups correspond with the states of the MDP from Fig.~\ref{fig:toy-mdp}, while different lines correspond to different initial values of $\beta(green)$ (a lighter color depicts a lower value). So there are three lines for each starting value of $\beta$,  one of each color. {\bf First row:} Termination-critic. {\bf Second row:} Naive reachability. We see that when the $\beta$-initialization is not too low, termination critic correctly concentrates termination on the attractor state and that state only, while the naive version saturates two of the states.} 
     \label{fig:dynamics}
 \end{figure*}

\section{Experiments}


We will now empirically support the following claims:
\begin{itemize}
    \item Our algorithm directs termination into a small number of frequently visited states;
    \item The resulting options are intuitively appealing and improve learning performance; and
    \item The predictability objective is related to planning performance and the resulting options improve both the objective and planning performance.
\end{itemize}

We will evaluate the latter two in the classical Four Rooms domain~\cite{sutton1999between} (see Figure~\ref{fig:rooms} in appendix). But before we begin, let us consider the learning dynamics of the algorithm on a small example. 

\subsection{Learning Dynamics}
\label{sec:learning-dynamics}

 The predictability criterion we have proposed is minimized by any termination scheme that deterministically terminates in a single state. Then, what solution do we expect for our algorithm to find? We hypothesize that the learning dynamics favor states that are visited more frequently, that is: are more reachable as measured by $\po$, up to the initial disbalance of $\beta$. In this section we validate this hypothesis on a small example.
 
  Consider the MDP in Fig.~\ref{fig:toy-mdp}. The green state is a potential attractor, with its attraction value being determined by a parameter $p$. Let the initial $\beta(blue) = \beta(red) = 0.5$. In this experiment, we investigate the resulting solution as a function of the initial $\beta(green)$ and $p$. Following the intuition above, we expect the algorithm to increase $\beta(green)$ more when these values are higher. We compare the results with another way of achieving a similar outcome, namely by simply using the marginal value $\margP$ as the objective ("naive reachability"). As expected, we find that the full predictability objective is necessary for concentrating in a {\em single} most-visited state. 
See Fig.~\ref{fig:dynamics}.
  
  We further performed an ablation on the two advantage terms that comprise our loss (the reachability advantage and the trajectory advantage); these results are given in Figure~\ref{fig:dynamics-full}. 
   We find that neither 
term in isolation is sufficient, or as effective as the full objective.

\vspace{-0.7em}
 \subsection{Option Discovery}
 \label{sec:experiments}

 We now ask how well the termination critic and the predictability objective are suited as a method for end-to-end option discovery. 
 As discussed, option discovery remains a challenging problem and end-to-end algorithms, such as the option-critic, 
 often have difficulties learning non-trivial or intuitively appealing results.
 
 We evaluate both ACTC and option-critic with deliberation cost (A2OC) on the Four Rooms domain using visual inputs (a top-down view of the maze passed through a convolutional neural network). The basic task is to navigate to a goal location, with zero per-step reward and a reward of $1$ at the goal. The goal changes randomly between a fixed set of eight locations every 20 episodes, with the option components being oblivious to the location, but the option-level value function being able to observe the location of the goal.

\begin{figure*}[t]
    \centering
    \includegraphics[width=\textwidth]{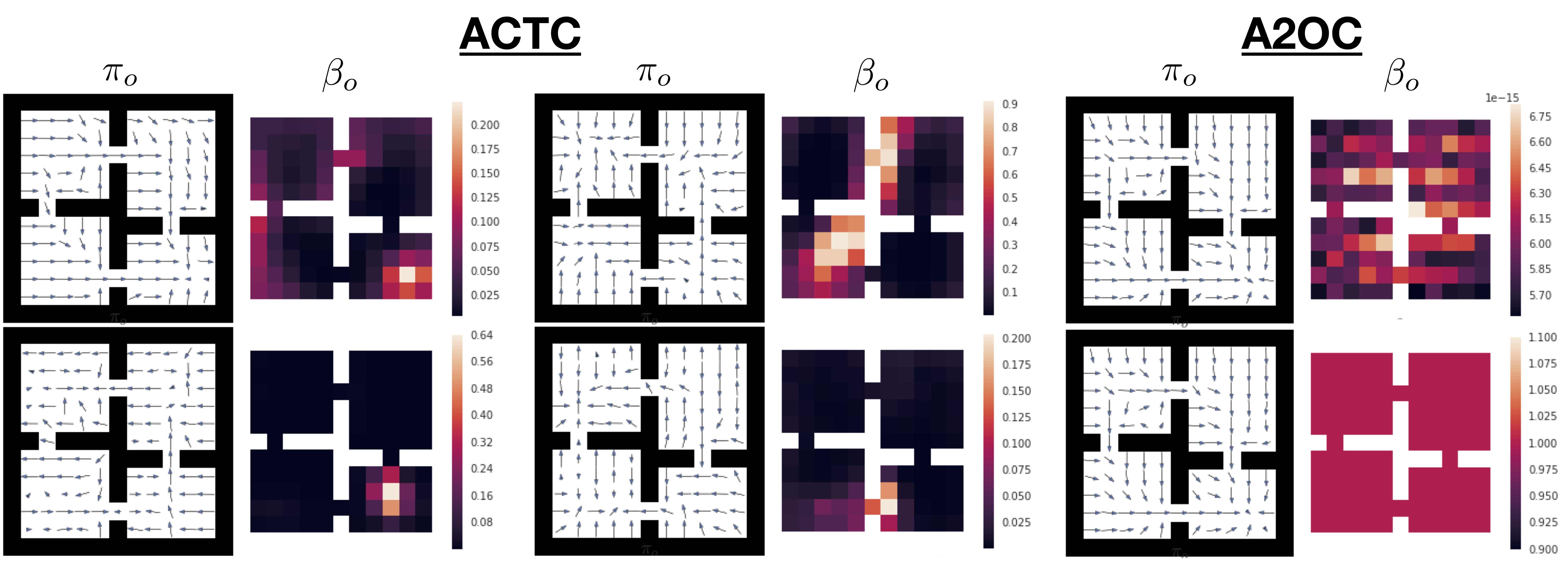}
    \caption{Example resulting options from ACTC (left) and A2OC (right). Each option is depicted via its policy and termination condition. ACTC concentrates termination probabilities around a small set of states while A2OC, with deliberation cost, tends to saturate on constant zero or constant one termination probability.}
    \label{fig:4opts}
\end{figure*}

\subsubsection{Architecture}

We build on the neural network architecture of A2OC with deliberation cost~\cite{harb2017waiting}, which in turn is almost identical to the network for A3C~\cite{mnih2016asynchronous}. Specifically, the A3C network outputs a policy $\pi$ and action-value function $Q$, whereas our network outputs $\{\pi^o\}_{o\in\O}$, and $\{Q(\cdot,o)\}_{o\in\O}$ given the input state. We use an additional network that takes in two input states, $x_s$ and $x_f$, and outputs $\{P^o(x_f | x_s)\}_{o\in\O}$, and $\{\bo(x_f)\}_{o\in\O}$, as well as some useful auxiliary predictions: a marginal $\{\po_\mu(x_f)\}_{o\in\O}$ and the update baseline $\{B^o\}_{o\in\O}$. This network uses a convolutional network with identical structure as used in A3C, that feeds into a single shared fully-connected hidden layer, followed by a specialized fully-connected layer for each of the outputs. 

\begin{figure}[t]
    \centering
    \includegraphics[width=.43\textwidth]{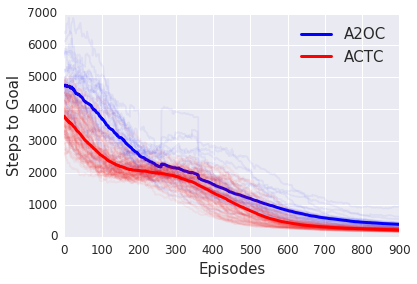}
    \caption{The learning performance of the two algorithms on the the Four Rooms task with switching goals. We plot the entire suite of hyperparameters, which for A2OC includes various deliberation costs, and for ACTC different learning rates for the $\beta$-network. We see ACTC exhibit better learning performance.}
    \label{fig:learning_curve}
\end{figure}

 \begin{figure}[t]
     \centering
    \includegraphics[width=.43\textwidth]{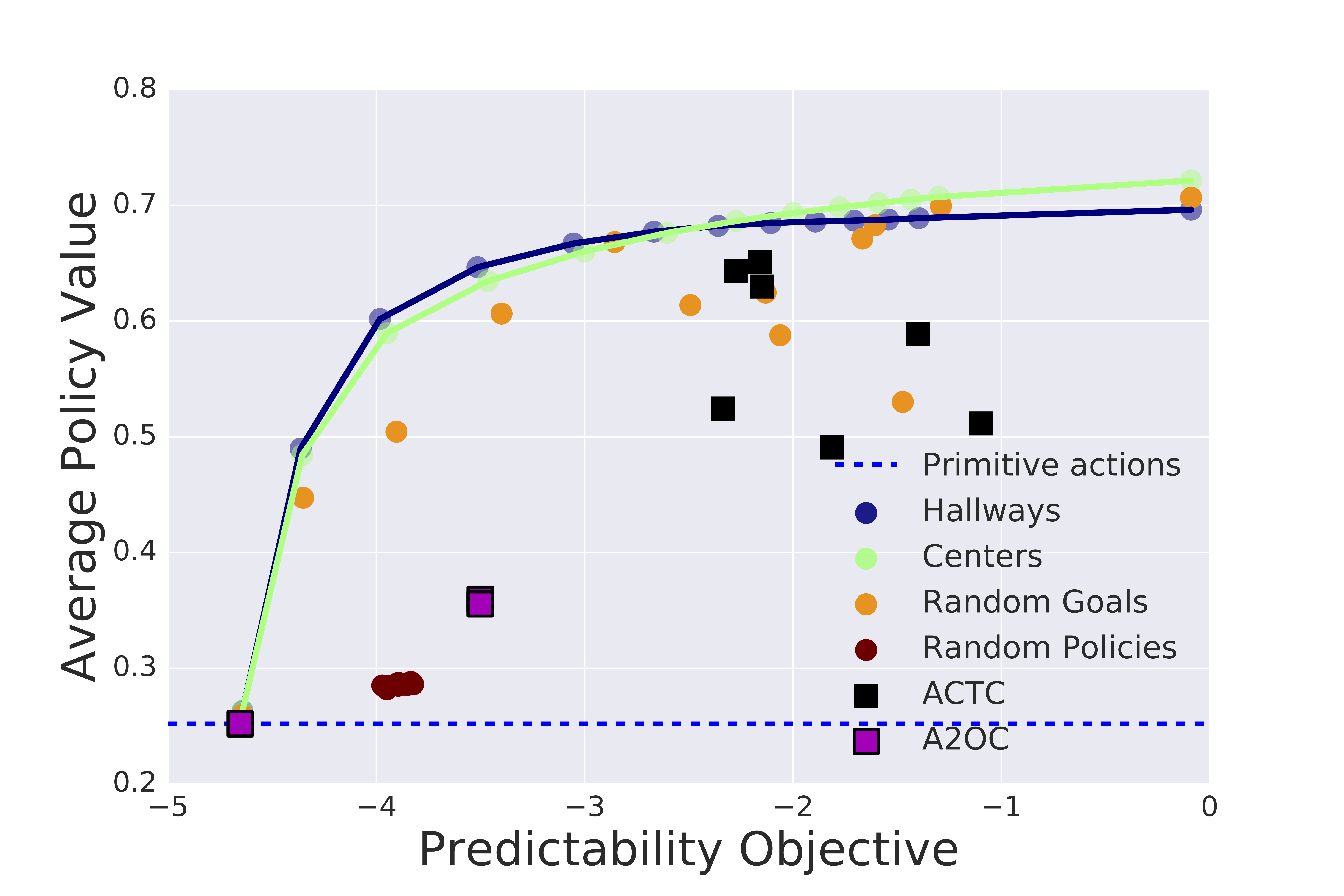}
    \caption{Investigating correlation between predictability and planning performance. Average policy value plotted against predictability objective (negative of the loss). A2OC options generalize poorly to unseen goals and have unpredictable terminations. ACTC optimizes the predictability objective leading to reusable options.}
     \label{fig:correlations}
 \end{figure}

\subsubsection{Learning and Planning in Four Rooms}

We first depict the options qualitatively with an example termination profile shown in Figure~\ref{fig:4opts}. We see that ACTC leads to tightly concentrated regions with high termination probability and low probability elsewhere, whereas A2OC even with deliberation cost tends to converge to trivial termination solutions. Although ACTC does not always converge to terminating in a single region, it leads to distinct options with characteristic behavior and termination profiles. 

Next, in Figure~\ref{fig:learning_curve} we compare the online learning performance between ACTC and A2OC with deliberation cost. The traces indicate separate hyper-parameter settings and seeds for each algorithm and the bold line gives their average. ACTC enjoys better performance throughout learning.

\subsection{Correlation with Planning Performance}
\label{sec:correlation-with-planning}

Finally,  we investigate the claim that more directed termination leads to improved planning performance. To this end, we generate various sets ($n = 4$) of goal-directed options in the Four Rooms domain by systematically varying the option-policy goal location and concentration of termination probability around the goal location. We evaluate these options, combined with primitive actions, by averaging the policy value during ten iterations of value iteration and all possible goal locations (see appendix for more details).

We compare this average policy value as a function of the predictability objective of each set of options in Figure~\ref{fig:correlations}. "Hallways" corresponds to one option for each hallway, "Centers" -- to the centers of each room, "Random Goals" -- to each option selecting a unique random goal location, and "Random Options" -- to both the policies and termination probabilities being uniformly random. We observe, as has previously been reported, that even random options improve planning over primitive actions alone (shown by the dashed-line). Additionally, we confirm that generally as the predictability objective increases towards zero the average policy value increases, corresponding to faster convergence of value iteration.

Finally, we plot (with square markers) the performance of options learned from the previous section using ACTC and A2OC with deliberation cost. Due to A2OC's option collapse, its advantage over primitive actions is small, while 
ACTC performs similarly to the better (more deterministic) random goal options.

\section{Related Work}

The idea of decomposing behavior into reusable components or abstractions has a long history in the reinforcement learning literature. One question that remains largely unanswered, however, is that of suitable criteria for identifying such abstractions. The option framework itself \cite{sutton1999between,bacon2017option} provides a computational model that allows the implementation of temporal abstractions but does in itself not provide an objective for option induction. This is addressed partially in \cite{harb2017waiting} where long-lasting options are explicitly encouraged. 

A popular means of encouraging specialization is via different forms of information hiding, either by shielding part of the policy from the task goal (e.g.\ \cite{heess2016learning}) or from the task reward (e.g. \cite{vezhnevets2017feudal}). \cite{frans2017meta} combine information hiding with meta-learning to learn options that are easy to reuse across tasks. 

Information-theoretic regularization terms that encourage mutual information between the option and features of the resulting trajectory (such as the final state) have been used to induce diverse options in an unsupervised or mixed setting (e.g.~\cite{gregor2016variational,florensa2017stochastic,eysenbach2018diversity}), and they can be combined with information hiding (e.g.~\cite{hausman2018learning}). Similar to our objective they can be seen to encourage predictable outcomes of options.    \cite{coreyes2018self} have recently proposed a model that directly encourages the predictability of trajectories associated with continuous option embeddings to facilitate mixed model-based and model-free control. 

Unlike our work, approaches described above consider options of fixed, predefined length. The problem of learning option boundaries has received some attention in the context of learning behavioral representations from demonstrations (e.g.~\cite{Daniel2016Probabilistic,Lioutikov2015Probabilistic,Fox2017Multi,Krishnan2017DDCO}). These approaches effectively learn a probabilistic model of trajectories and can thus be seen to perform trajectory compression.

Finally, another class of approaches that is related to our overall intuition is one that seeks to identify "bottleneck" states and construct goal-directed options to reach them. There are many definitions of bottlenecks, but they are generally understood to be states that connect different parts of an environment, and are hence visited more often by successful trajectories. Such states can be identified through heuristics related to the pattern of state visitation~\cite{mcgovern2001,stolle2003} or by looking at between-ness centrality measures on the state-transition graphs. Our objective is based on a similar motivation of finding a small number of states that give rise to a compressed high-level decision problem which is easy to solve. However, our algorithm is very different (and cheaper computationally).

\section{Discussion}


We have presented a novel option-discovery criterion that uses predictability as a means  of regularizing the complexity of behavior learned by individual options. We have applied it to learning meaningful  {\em termination conditions} for options, and have demonstrated its ability to induce options that are non-trivial and useful. 

Optimization of the criterion is achieved by a novel policy gradient formulation that relies on learned option models. In our implementation we choose to decouple  the reward optimization from the problem of learning where to terminate. This particular choice allowed us to study the effects of meaningful termination in isolation. We saw that even if the option policies optimize the same reward objective, non-trivial terminations prevent option collapse onto the same policy.

This work has focused entirely on  goal-directed options, whose purpose is to reach a certain part of the state space.  There is another class, often referred to as skills, which are in a sense complementary, aiming to abstract behaviors that apply {\em anywhere} in the state space. One exciting direction for future work is to study the relation between the two and to design option induction criteria that can interpolate between different regimes.

\bibliographystyle{apalike}
\bibliography{bibliography}

\clearpage

\appendix

\section{The Option Transition Process}
\label{sec:transition-process}
It will be convenient to consider the option transition process:
\begin{align}
\paug{0}(x_f|x_s) & = \Pr(x_{t} = x_f|x_t = x_s) = \indic{x_s=x_f} \notag \\
\paug{1}(x_f|x_s) & = \Pr(x_{t+1} = x_f|x_t = x_s) \notag \\
& = (1-\bo(x_s)) p^{\pio}(x_f|x_s)  \notag \\ 
& \ldots  \notag \\
\paug{k}(x_f|x_s) & = \Pr(x_{t+k} = x_f|x_t=x_s) \notag \\
& = \sum_x \paug{1}(x|x_s) \paug{k-1}(x_f|x)  \notag 
\end{align}
We can then rewrite $\po$ from \eqref{eq:po} as: 
\begin{align}
\po(x_f|x_s) & = \bo(x_f) \left( \paug{0}(x_f | x_s) + \paug{1}(x_f | x_s) + \ldots \right)  \notag \\
& = \bo(x_f) \sum_{k=0}^\infty \paug{k}(x_f | x_s)  \label{eq:po-as-sum}
\end{align}

\section{Omitted Proofs}

\subsection{Proof of Theorem~\ref{thm:po-gradient}}
\begin{proof}
We have:
\begin{align}
& \dP{x_f|x_s} \notag \\
&= \dbeta{x_s}\indic{x_f=x_s} + \deriv(1-\bopar(x_s))\sum_{x}\ppio(x|x_s) {\po(x_f|x)} \notag \\
& = \dbeta{x_s}\indic{x_f=x_s} + \sum_{x}\ppio(x|x_s) \Big(\dP{x_f|x} 
\notag \\ & \qquad 
- \deriv\Big(\bopar(x_s)\po(x_f|x) \Big)\Big) \notag \\
& = \dbeta{x_s}\indic{x_f=x_s} + \sum_{x}\ppio(x|x_s) \Big(\dP{x_f|x} 
\notag \\ & \qquad -\dbeta{x_s}\po(x_f|x) - \bopar(x_s)\dP{x_f|x} \Big) \notag \\
& = \dbeta{x_s}\Big(\indic{x_f=x_s} - \sum_{x}\ppio(x|x_s) \po(x_f|x) \Big) 
\notag \\ & \qquad  
+ (1-\bopar(x_s)) \sum_{x}\ppio(x|x_s) \dP{x_f|x}. \label{eq:dp}
\end{align}
And so what we have is a  $(1-\bopar(x_i))$-discounted value function, whose reward is  $\dbeta{x_i}\rox{x_i}{x_f} $, where 
\begin{align}
\rox{x_i}{x_f} & = \indic{x_f=x_i} - \sum_{x_{i+1}}p^{\pio}(x_{i+1}|x_i){\po(x_f|x_{i+1})} \notag 
\end{align} 
Now, from Eq.~\eqref{eq:po} and if $\bo(x) \neq 1$, we have:
\begin{align}
    \sum_{x}\ppio(x|x_s) &\po(x_f|x)  = \frac{ \po(x_f|x_s)  -  \bo(x_s) \indic{x_f=x_s}}{1 - \bo(x_s)} \notag \\
    \rox{x_s}{x_f} & =  \indic{x_f=x_s} - \frac{ \po(x_f|x_s)  -  \bo(x_s) \indic{x_f=x_s}}{1 - \bo(x_s)} \notag \\
    & = \frac{\indic{x_f=x_s} - \po(x_f|x_s)} {1-\bo(x_s)}
\end{align}

Using this notation, and recalling the transition process from Eq.~\eqref{eq:po-as-sum}, we can rewrite \eqref{eq:dp} as:
\begin{align}
& \dP{x_f|x_s} \notag \\
& = \dbeta{x_s}\rox{x_s}{x_f} + \sum_{x} \paug{1}(x|x_s) \dP{x_f|x} \notag \\
& =  \sum_{x} \sum_{k=0}^\infty \paug{k}(x|x_s) \dbeta{x}\rox{x}{x_f} \notag \\
& = \sum_{x} \frac{\po(x|x_s)}{\bo(x)} \dbeta{x} \rox{x}{x_f} \notag \\
& = \sum_{x} \po(x|x_s) \dlogbeta{x}  \rox{x}{x_f} \notag 
\end{align}
Where the third equality follows from \eqref{eq:po-as-sum} and requires for $\bo(x)$ to not be $0$.


\end{proof}

\subsection{Proof of Proposition~\ref{prop:criteria-po}}
\begin{proof}
Let $\Pr(x|o)$ denote the probability of a state $x$ being terminal for an option $o$. By definition of entropy we have:
\begin{align*}
H(X_f|o) & = - \sum_{x_f}\Pr(x_f|o) \log\Pr(x_f|o) \\
& = -\sum_{x_f} \sum_{x_s} \Pr(x_s|o) \Pr(x_f|x_s,o)\\
& \qquad\qquad \times \log\sum_{x_s} \Pr(x_s|o) \Pr(x_f|x_s,o) \\
& = -\sum_{x_f} \sum_{x_s} \dmu{x_s} \po(x_f|x_s)  \\
& \qquad\qquad \times \log \underbrace{\sum_{y_s} \dmu{y_s}\po(x_f|y_s)}_{\mbox{marginal } \margP(x_f)} \\
& = -\sum_{x_s} \dmu{x_s} \sum_{x_f}\po(x_f|x_s) \log\margP(x_f)
\end{align*}
\end{proof}

\subsection{Proof of Theorem~\ref{thm:main-result}}
\begin{proof}
\begin{align*}
    & \deriv J(\po)  = -\deriv \underbrace{\sum_{x_s} \dmu{x_s}}_{\E_{x_s}} \sum_{x_f} \po(x_f|x_s)  \log \margP(x_f)\\
    & = -\E_{x_s} \Big[\sum_{x_f} \Big(\deriv\po(x_f|x_s) \log \margP(x_f) \\
    & \qquad \qquad + \po(x_f|x_s) \frac{\deriv\margP(x_f)}{\margP(x_f)} \Big)\Big] \\
    & = -\E_{x_s}\Big[\sum_{x_f} \Big(\sum_{x} \po(x|x_s) r^o_{x_f}(x) \dlogbeta{x} \log \margP(x_f) \\
    & \quad + \frac{\po(x_f|x_s)}{\margP(x_f)} \underbrace{ \sum_{y_s}\dmu{y_s} \sum_{x} \po(x|y_s)}_{\sum_x \margP(x)}r^o_{x_f}(x) \dlogbeta{x}  \Big) \Big] \\
    & = -\E_{x_s} \Big[ \sum_{x} \po(x|x_s)\dlogbeta{x}\sum_{x_f} r^o_{x_f}(x) \\
    & \quad \times \Big( \log \margP(x_f) + \frac{\po(x_f|x_s)}{\margP(x_f)}\frac{\margP(x)}{\po(x|x_s)} \Big) \Big] \\
    & = -\E_{x_s} \Big[\sum_{x} \po(x|x_s) \frac{\dbeta{x}}{\bo(x)} \sum_{x_f} \frac{\indic{x_f=x} - \po(x_f|x)}{1-\bo(x)} \\
     & \quad \times \Big( \log \margP(x_f) + \frac{\po(x_f|x_s)}{\margP(x_f)}\frac{\margP(x)}{\po(x|x_s)} \Big) \Big]\\
    & = -\underbrace{\sum_{x_s} \dmu{x_s}}_{\mbox{\small sample}} \underbrace{\sum_{x} \frac{\po(x|x_s)}{\bo(x)}}_{\mbox{\small sample (continuation)}} \frac{\dbeta{x}}{1-\bo(x)} \\
    & \quad \times \Big[ \Big( \log \margP(x) + 1 \Big)\\
    & \quad - \underbrace{\sum_{x_f} \po(x_f|x)}_{\mbox{\small sample}} \Big( \log \margP(x_f) + \frac{\po(x_f|x_s)\margP(x)}{\margP(x_f)\po(x|x_s)} \Big)  \Big]
\end{align*}
Sampling the highlighted expectations, and noting that if $\ell$ are the logits of $\bo$,
\[\deriv\ell_{\bo}(x) = \frac{\dbeta{x}}{\bo(x)(1-\bo(x))},\]
we have our result.

\end{proof}

\section{Correlation with Planning Performance}
\label{app:correlation}

\begin{figure}
    \centering
    \includegraphics[width=0.5\textwidth]{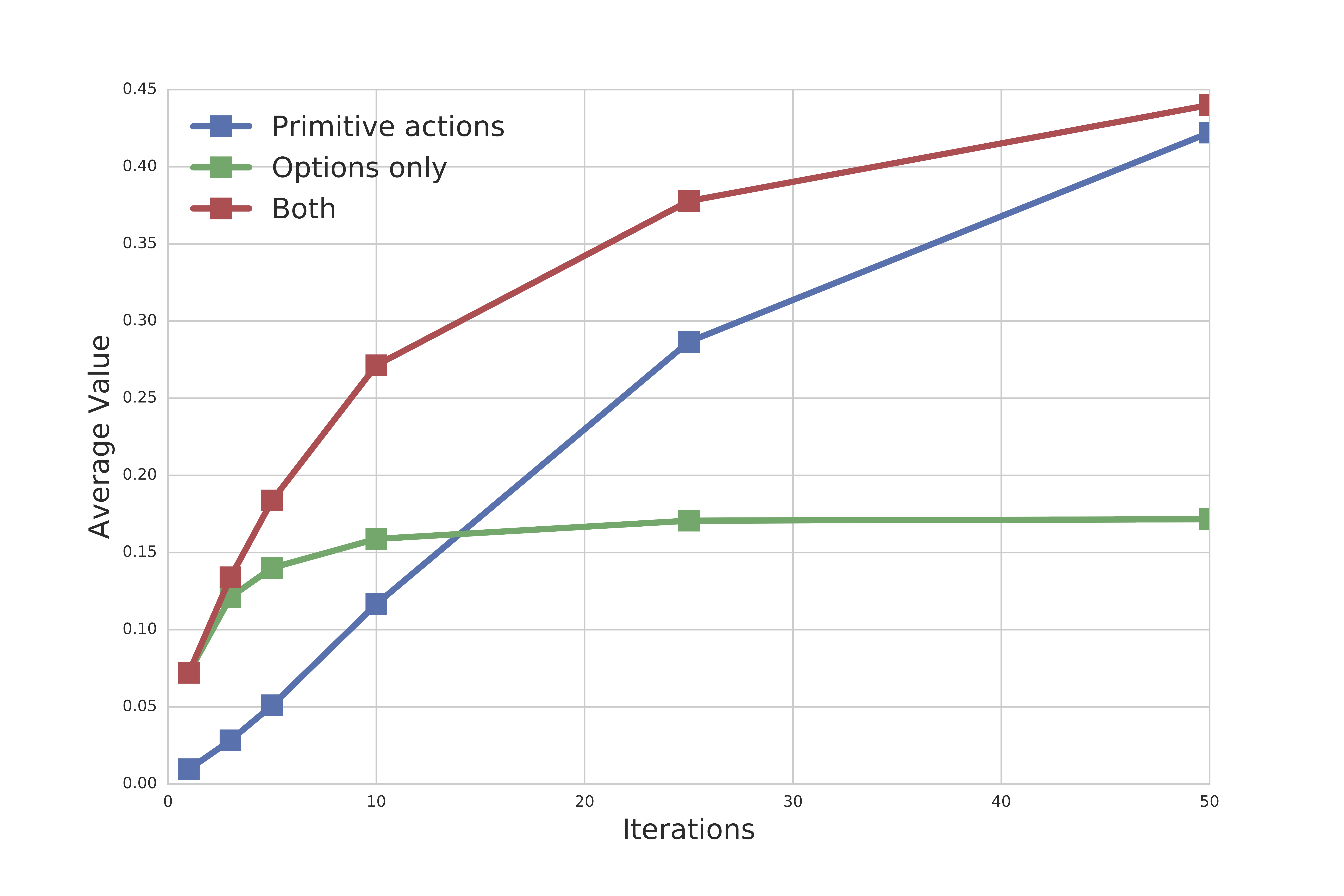}
    \caption{Example of planning performance using options, primitive actions, and both options and primitive actions.
    }
    \label{fig:planning_example}
\end{figure}

The policies considered in these experiments consist of some set of four options combined with the set of primitive actions. Planning performance, for a single goal-directed task, is evaluated as the average policy value over all states at the end of each of ten iterations of value iteration. Consider Figure~\ref{fig:planning_example} which shows the value iteration performance curve for a single task, comparing policies of primitive actions, options, and their combination. The planning performance is the average of this curve for ten iterations, further averaged over all possible goal-directed tasks in Four Rooms. This measures how quickly value iteration, using this set of option policies and terminations, is able to plan.

\begin{figure}[t]
    \centering
    \includegraphics[scale=0.6]{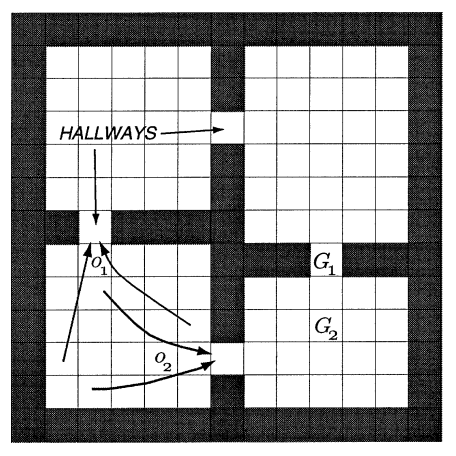}
    \caption{The Four Rooms domain map. 
    }
    \label{fig:rooms}
\end{figure}

\section{Learning Dynamics}
\label{app:learning-dynamics}

\begin{figure*}
     \centering
     \includegraphics[scale=0.4]{tc-dynamics.png}
     \includegraphics[scale=0.4]{just-po-dynamics.png}
     \includegraphics[scale=0.4]{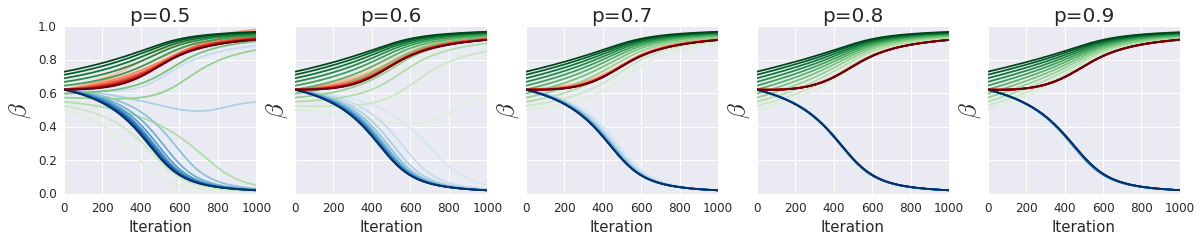}
     \includegraphics[scale=0.4]{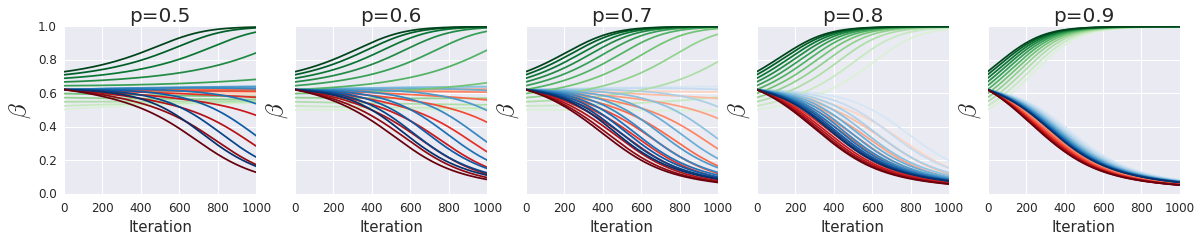}
     \caption{Learning dynamics. The color groups correspond with the states of the MDP from Fig.~\ref{fig:toy-mdp}, while different lines correspond to different initial values of $\beta(green)$ (a lighter color depicts a lower value), {\bf First row:} Termination-critic. {\bf Second row:} Naive reachability. {\bf Third row:} Only termination score advantage. {\bf Fourth row:} Only relative termination advantage. We see that when the $\beta$-initialization is not too low, termination critic correctly concentrates termination on the attractor state and that state only, while the naive version saturates two of the states. The two ablations show the reachability advantage having similar behavior to naive reachability, while the trajectory advantage is not concentrating enough when the attraction values are low. 
     }
     \label{fig:dynamics-full}
 \end{figure*}

Fig.~\ref{fig:dynamics-full} further studies the learning dynamics induced by the different components of the algorithm. We compare the previous two variants from Fig.~\ref{fig:dynamics} with only including the reachability advantage term (Row 3), and only including the trajectory advantage term (Row 4). The former does not focus on a single state, while the latter does not concentrate at all for many values of $\beta$.

\end{document}